\documentclass{sig-alternate}
\usepackage{booktabs}
\usepackage{color}
\usepackage{url}
\usepackage{graphics}
\usepackage[usenames,dvipsnames]{xcolor}

\newcommand{\hide}[1]{}
\newcommand{\todo}[1]{\textcolor{red}{[\textit{#1}]}}

\newcommand{\proj}{M3}

\begin{document}
%
% --- Author Metadata here ---
\CopyrightYear{2016} 
\setcopyright{rightsretained} 
\conferenceinfo{SIGMOD/PODS'16}{June 26 - July 01, 2016, San Francisco, CA, USA} 
\isbn{978-1-4503-3531-7/16/06}
\doi{http://dx.doi.org/10.1145/2882903.2914830}
%\CopyrightYear{2007} % Allows default copyright year (20XX) to be over-ridden - IF NEED BE.
%\crdata{0-12345-67-8/90/01}  % Allows default copyright data (0-89791-88-6/97/05) to be over-ridden - IF NEED BE.
% --- End of Author Metadata ---

\title{{\ttlit \proj{}}: Scaling Up Machine Learning via Memory Mapping}

\numberofauthors{2} %  in this sample file, there are a 
\author{
\alignauthor
Dezhi Fang\\
       \affaddr{College of Computing}\\
       \affaddr{Georgia Institute of Technology}\\
       \email{dezhifang@gatech.edu}
% 2nd. author
\alignauthor
Duen Horng Chau\\
       \affaddr{College of Computing}\\
       \affaddr{Georgia Institute of Technology}\\
       \email{polo@gatech.edu}
}
% There's nothing stopping you putting the seventh, eighth, etc.
% author on the opening page (as the 'third row') but we ask,
% for aesthetic reasons that you place these 'additional authors'
% in the \additional authors block, viz.
% Just remember to make sure that the TOTAL number of authors
% is the number that will appear on the first page PLUS the
% number that will appear in the \additionalauthors section.
\maketitle

\begin{abstract}

%\todo{shorten significantly}
% The ability to compute large out-of-core data sets with machine learning methods has become the foundation of modern data science. 
% Most methods available today focus on using distributed systems to achieve high performances and scalability.
% However, these methods often suffer from considerable overhead inherent to the complicated nature of distributed systems.
% Such methods often sacrifice overall performance for their pursuit in scalability.
To process data that do not fit in RAM, conventional wisdom would suggest using distributed approaches.
However, recent research has demonstrated virtual memory's strong potential in scaling up graph mining algorithms on a single machine.
%Recent research has demonstrated virtual memory's strong potential in scaling up graph mining algorithms on a single machine. 
We propose to use a similar approach for general machine learning.
% propose a minimalistic approach towards scaling up machine learning by using using a similar approach. 
We contribute: 
(1) our latest finding that memory mapping is also a feasible technique for scaling up general machine learning algorithms like logistic regression and k-means, when data fits in or exceeds RAM (we tested datasets up to 190GB);
(2) an approach, called \proj{}, that enables existing machine learning algorithms to work with out-of-core datasets through memory mapping, 
% with minimal changes to the code, 
achieving a speed that is significantly faster than a 4-instance Spark cluster, and comparable to an 8-instance cluster.
% , and are often more than twice as fast as a 4-instance cluster.
% can easily adapt to take advantage of our technique;
% (3) preliminary experiments on the model's performance in comparison to Spark.
\end{abstract}


% \keywords{Machine Learning; Virtual Memory}

% \begin{figure}[t]
% \centering
% \epsfig{file=scalesangria.png, width=3in}
% \caption{Comparing Running time of M3 to the Size of Data Set}
% \label{fig:scalability}
% \end{figure}

\section{Introduction}
Leveraging virtual memory to extend algorithms for out-of-core data has  received increasing attention in data analytics communities. 
Recent research demonstrated virtual memory's strong potential to scale up graph algorithms on a single PC \cite{mcsherry2015scalability,lin2014mmap}.
Available on almost all modern platforms, virtual memory based approaches are straight forward to implement and to use, and can handle graphs with as many as 6 billion edges \cite{lin2014mmap}.
Some single-thread implementations on a PC can even outperform popular distributed systems like Spark (128 cores) \cite{mcsherry2015scalability}.
% , which may require significant expertise on refactoring existing code base as well as tuning and managing the clusters.
% It is shown that \textit{MMap} \cite{lin2014mmap} can handle graph with more than 6 billion edges.
Memory mapping a dataset into a machine's virtual memory space allows the dataset to be treated identically as an in-memory dataset.
% The memory management of such out-of-core, memory-mapped dataset is deferred to the operating system (OS).
The algorithm developer no longer needs to explicitly determine how to partition the (large) dataset, nor manage which partitions should be loaded into RAM, or unloaded from it.
The OS performs similar actions on the developer's behalf, through paging the dataset in and out of RAM, via highly optimized OS-level operations.

% Virtual memory, an abstraction provided by the operating system, maps memory addresses to a variety of physical locations including RAM, hard drives and network sockets \cite{deri2004improving}, and has the advantage of transparency to algorithm implementations, i.e., developers can defer the memory management of out-of-core data to the operating system (OS).

% A popular alternative solution is to use distributed systems, such as Spark \cite{zaharia2010spark}.
% , are proven to be able to scale up machine learning algorithms efficiently.
% However, such systems may require significant expertise on refactoring existing code base as well as tuning and managing the clusters.
%despite having great scalability, 
%As suggested by previous works in graph mining algorithms, virtual memory can be applied to scale up those algorithms with relatively low overhead.

\begin{figure}[t]
    \centering
    \includegraphics[width=0.48\textwidth]{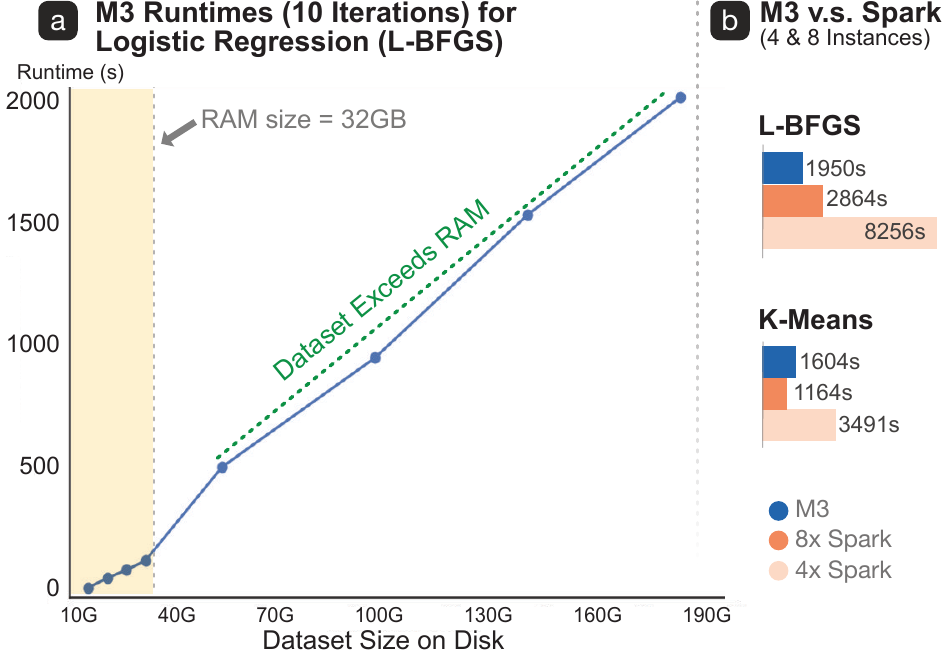}
    \caption{a: \proj{} runtime scales linearly with data size, when data fits in or exceeds RAM. 
    b: \proj{}'s speed (one PC) comparable to 8-instance Spark (orange), and significantly faster than 4-instance Spark (light orange).}
    \label{fig:scalability}
\end{figure}

\section{Scaling Up using \proj{}}
\hide{
\todo{1. what are we doing? Fit dataset (which part of the data) into virtual memory. What function do we need to call; which parts of existing code are we modifying}

\todo{2. potential benefits of using \proj{}/virtual memory: easiser to program; potentiall less overhead because we defer memory management to OS, etc.}

\todo{3. why it would work, COST paper, memory access pattern, read-ahead, LRU, to exploit locality, e.g. sequential scan works, SGD because not fit in main memory, \proj{} does not fit in main memory, use cite L-BFGS converges faster, random access is bad}
}
% Contrary to established methods of employing distributed systems, we are exploiting the transparency provided by virtual memory. 
% Our investigation takes a first major step in assessing whether virtual memory is a suitable 
As existing works focused on graph algorithms such as PageRank and finding connected components, 
we are investigating whether a similar virtual memory based approach can generalize to machine learning algorithms at large.
% as a fundamental, radical scale-up approach that is also easy to apply. 

Inspired by prior works on graph computation, 
our \proj{}\footnote{\proj{} stands for \underline{M}achine Learning via \underline{M}emory \underline{M}apping.} approach uses memory mapping to amplify a single machine's capability to process large amounts of data for machine learning algorithms. 
%After memory mapping a dataset, programming can be performed on it as if that data were in-memory. 
As memory mapping a dataset allows it to be   treated identically as an in-memory dataset, \proj{} is a transparent scale-up strategy that developers can easily apply, requiring minimal modifications to existing code. 
For example, Table \ref{table:code} shows that 
with only minimal code changes and a trivial helper function, existing algorithm implementation can easily handle much larger, memory-mapped datasets.

% only one line of modification to the main program with a trivial helper function that handles the data loading and memory allocation.
% with only minimal code changes and a trivial helper function, existing algorithm implementation can easily handle much larger, memory-mapped datasets.
% that handles the data loading and memory allocation.
% that using in-memory data structure, enabling it to work with much larger memory-mapped data.

Modern 64bit machines have address spaces large enough to fit large datasets into (up to 1024PB). 
Because the operating system has access to a variety of internal statistics on how the mapped data is being used, the access to such data can be further optimized by the operating system via methods including least recent used caching and read-ahead to achieve efficiency \cite{boyd2010analysis}. 

To test the feasibility of \proj{}, we minimally modified mlpack, an efficient machine learning library written in C++ \cite{mlpack2013}. Memory mapping can be easily applied to other languages and algorithm libraries.

% \usepackage{booktabs}
\begin{table}[t]
\small
\centering

\begin{tabular}{@{}l|l@{}}
\toprule
\textbf{Original}   & \textbf{\proj{}}                                                                                                                  \\ \midrule
\texttt{Mat data;} & \begin{tabular}[c]{@{}l@{}}\texttt{double *m = mmapAlloc(file, rows * cols);}\\ \texttt{Mat data(m, rows, cols);}\end{tabular} \\ \bottomrule
\end{tabular}
\caption{\proj{} introduces minimal changes to code  originally using in-memory data structure, enabling it to work with much larger memory-mapped data.}
\label{table:code}
\end{table}

% Data sets are preprocessed into packed binaries for easier loading purposes. 
%The prepossessing does not exceed parsing integers and floating point numbers into their binary representations.
%\todo{Our approach is easy to apply by design, requiring minimal modification to existing code. For example... Figure 1 shows only one line of modification to the main program with a trivial helper function of loading the data set into the virtual memory.}

%\andy{Revise?}Our investigation of scaling up machine learning via memory mapping...
% 
%As shown in Table 1, \proj{} requires trivial modifications to existing machine learning algorithms.
%Our preliminary experiments suggest that ...

%\todo{don't mention SGD since you aren't reporting its results}
%Stochastic Gradient Descent method is often used when the data cannot be fit in memory. 
%However, with \proj{}, methods that require the entire dataset to be in-memory, such as L-BFGS can be applied. L-BFGS is especially suitable to \proj{} because it has a mainly sequential scan memory access pattern, when the kernel's optimizations can be applied to exploit the locality. 

% In a typical program of M3, data sets are first loaded as matrices into memory, and then, further matrix manipulation techniques can be applied as normal. The implementation of machine learning algorithms can be heavily optimized as if such data fit into memory to achieve better efficiency.

\section{Experiments}
Our current evaluation focuses on:
(1) understanding of how \proj{} scales with increasing data sizes; and
(2) how \proj{} compares with distributed systems such as Spark, as prior work suggested the possibility that a single machine can outperform a computer cluster \cite{mcsherry2015scalability}.

% We performed preliminary evaluation of M3 to better understand how it scales with increasing data size.
% We are also curious about how M3 compares with distributed systems such as Spark, as prior work suggested the possibility that single machines can outperform computer clusters \cite{mcsherry2015scalability}.

% \subsection{Comparison with Spark}
% In order to normalize the performance of machines used in experiments, 

\textbf{Experiment Setup.}
All tests with \proj{} are conducted on a desktop computer with Intel i7-4770K quad-core CPU at 3.50GHz (8 hyperthreads), 4$\times$8GB RAM, 1TB SSD of OCZ RevoDrive 350.
% and 2$\times$3TB WD 7200RPM hard disk.
We used Amazon EC2 \texttt{m3.2xlarge} instances for Spark experiments.
% are both conducted on \texttt{m3.2xlarge} machines. 
Each instance has 8 vCPUs (hyperthreads of an Intel Xeon core) with 30GB memory and 2$\times$80GB SSDs. 
%Intel Xeon E5-2670 CPUs
The Spark clusters are created by Amazon Elastic MapReduce and the datasets are stored on the cluster's HDFS.

\textbf{Dataset.} We used the \textit{Infimnist}\footnote{\url{http://leon.bottou.org/projects/infimnist}} dataset, an infinite supply of digit images (0--9) derived from the well-known MNIST dataset using pseudo-random deformations and translations. 
% \textit{Infimnist} enables us to easily generate realistic datasets of arbitrary sizes. 
Each image is 28$\times$28 pixel grayscale image (784 features; each image is 6272 bytes). 
We generated up to 32M images, whose dense data matrix representation contains 23.5 billion entries,
 amounting to 190GB.
Smaller datasets are subsets of the full 32M images.
% Each pixel is stored as a \textit{double} on disk. 
% Therefore, each image takes about 6.3 KB of storage. 

\textbf{Algorithms Evaluated.} 
We selected \textit{logistic regression} (L-BFGS for optimization) and k-means, 
since they are commonly used classification and clustering algorithms.\footnote{We are primarily interested in runtimes, so we did not perform image pre-processing.}
% We examined two methods of logistic regression: L-BFGS and gradient descent. 
% We used k-means++ in our experiments with k-means clustering.

\hide{
\begin{table}[]
\centering
\begin{tabular}{@{}llll@{}}
\toprule
Setup   & L-BFGS & Gradient Descent & Randomized SGD    \\ \midrule
MMap    & 5232s  & 616s       & 2076s             \\
4xSpark & 5500s  & 605.6s     & N/A (Job Failure) \\
8xSpark & 200s   & 100+s      & 100+s             \\ \bottomrule
\end{tabular}
\caption{Comparing \proj{} to Spark}
\label{my-label}
\end{table}
}

% To demonstrate M3's scalability, we randomly generated 20 dense matrices, each with rows ranging from 1 million to 20 million and 500 columns. The files' sizes, in csv format, ranges from 9GB to 177GB. 
% Experiments in this section is conducted on a desktop machine with Intel i7-4770K quad-core CPU at 3.50GHz, 4$\times$8GB RAM and 1TB SSD of Samsung 840 EVO-Series.

\subsection{Key Findings \& Implications}

\noindent \textbf{1. \proj{} scales linearly when data fits in RAM and when out-of-core,} for logistic regression (Figure \ref{fig:scalability}a).
The dotted vertical line in the figure indicates RAM size (32GB).
\proj{}'s runtime scales linearly both when the dataset fits in RAM (yellow region in Fig. \ref{fig:scalability}a), and when it exceeds RAM (green dotted line), at a higher scaling constant, as expected.

% However, once the data passes the size of the RAM, M3 continues to scale linearly.
Looking at \proj{}'s resource utilization, we saw that \proj{} is I/O bound: disk I/O was 100\% utilized while  CPU was only utilized at around 13\%. 
This suggests strong potential for \proj{} reaching even higher speed if we use faster disks, or configurations such as RAID 0.

\noindent \textbf{2. \proj{}'s speed (one PC) is comparable to 8-instance Spark and significantly faster than 4-instance Spark} for logistic regression (L-BFGS) and k-means (Figure \ref{fig:scalability}b). 
This result echoes prior works focusing on graph computation that suggests cluster may not be necessary for  moderately-sized datasets \cite{mcsherry2015scalability,lin2014mmap}.
Our result extends those findings to two general machine learning methods.

For \textit{logistic regression} (with 10 iterations of L-BFGS), \proj{} is about 30\% faster than 8-instance Spark.
4-instance Spark's runtime was 4.2X that of \proj{}.
For \textit{k-means} (10 iterations, 5 clusters), \proj{} ran at a speed comparable to 8-instance Spark (1.37X), and more than twice as fast as 4-instance Spark.

Certainly, using more Spark instances will increase speed, but that may also incur additional overhead (e.g., communication between nodes).
% , but that will also increases cost (e.g., AWS credit).
Here, we showed that for moderately-sized datasets, single-machine approaches like \proj{} can be attractive alternatives to distributed approaches.
% , avoiding the overhead introduced by a Spark cluster.
% For datasets that are much larger, distributed systems will be necessary.

\section{Conclusions \& Ongoing Work}
% There is no one-size fits all solution to solving the problem of scaling up machine learning. 
We are taking a first major step in assessing the feasibility of using virtual memory as a fundamental, alternative way to scale up machine learning algorithms.
\proj{} adds an interesting perspective to existing solutions primarily based on distributed systems.
% to solving the scaling problem: by using abstractions provided by the operating system.

We contribute:
(1) our latest finding that memory mapping could become a feasible technique for scaling up general machine learning algorithms when the dataset exceeds RAM;
% as an easy-to-work-with layer of abstraction for machine learning algorithms. The algorithms can be implemented as if the data fit into memory; 
(2) \proj{}, an easy-to-apply approach that enables existing machine learning implementations to work with out-of-core datasets;
% an analysis that memory mapping has low scaling overhead. This analysis suggests that memory mapping can be linearly scaled even higher; and
(3) our observations that \proj{} on a PC can achieve a 
speed that is significantly faster than a 4-instance Spark cluster, and comparable to an 8-instance cluster.

% has run-time that is competitive to a moderately-sized Spark cluster. 

% \noindent \textbf{3. Scalability results similar to the above applies to k-means.} 
% Figure \ref{fig:scalability} shows the \proj{}'s runtime comparison of k-means (10 iterations, 5 clusters) between \proj{} and Spark.
%\todo{... and refer to table/chart}
% This preliminary result suggests that virtual memory has the potential to generalize to more machine learning algorithms. 
% Here, we only tested logistic regression and k-means. 
We will extend our \proj{} approach to a wide range of machine learning (including online learning) and data mining algorithms. 
%The algorithms we have studied so far require mostly sequential scans, with occasional random accesses. 
We plan to extensively study the memory access patterns and locality of algorithms (e.g., sequential scans vs random access) to better understand how they they affect performance.
% For example, our preliminary study shows that \proj{}'s speed can suffer if an algorithm requires a lot of random accesses for datasets that vastly exceed RAM size. 
We plan to develop mathematical models and systematic approaches to profile and predict algorithm performance and energy usage based on extensive evaluations across platforms, datasets, and languages. 

% Further investigation into other machine learning algorithms that can take better advantage of locality is the logical next step in the research. 

%We will also investigate the possibility of using \proj{} on distributed systems, further utilizing abstractions like shared memory access offered by Lustre \cite{braam2002lustre}.

\section{Acknowledgments}
We thank Dr. Ryan Curtin and Minsuk (Brian) Kahng for their feedback. This work was supported by NSF grants IIS-1217559, TWC-1526254, IIS-1563816, and by gifts from Google, Symantec, Yahoo, eBay, Intel, Amazon,  LogicBlox.

\bibliographystyle{abbrv}
\bibliography{sigproc}

\begin{thebibliography}{1}

\bibitem{boyd2010analysis}
S.~Boyd-Wickizer, A.~T. Clements, Y.~Mao, A.~Pesterev, M.~F. Kaashoek,
  R.~Morris, N.~Zeldovich, et~al.
\newblock An analysis of linux scalability to many cores.
\newblock In {\em OSDI}, volume~10, pages 86--93, 2010.

\bibitem{mlpack2013}
R.~R. Curtin, J.~R. Cline, N.~P. Slagle, W.~B. March, P.~Ram, N.~A. Mehta, and
  A.~G. Gray.
\newblock {mlpack}: A scalable {C++} machine learning library.
\newblock {\em Journal of Machine Learning Research}, 14:801--805, 2013.

\bibitem{lin2014mmap}
Z.~Lin, M.~Kahng, K.~M. Sabrin, D.~H.~P. Chau, H.~Lee, and U.~Kang.
\newblock Mmap: Fast billion-scale graph computation on a pc via memory
  mapping.
\newblock In {\em Big Data (Big Data), 2014 IEEE International Conference on},
  pages 159--164. IEEE, 2014.

\bibitem{mcsherry2015scalability}
F.~McSherry, M.~Isard, and D.~G. Murray.
\newblock Scalability! but at what cost?
\newblock In {\em 15th Workshop on Hot Topics in Operating Systems (HotOS XV)},
  2015.

\end{thebibliography}
\end{document}